\def\year{2015}
\documentclass[letterpaper]{article}
\usepackage{aaai}
\usepackage{times}
\usepackage{helvet}
\usepackage{courier}

\usepackage[usenames,dvipsnames]{xcolor}
\usepackage{amsfonts,eucal,amsbsy,mathtools}
\usepackage{bm}
\usepackage{rotating}
\usepackage{tabularx}
\usepackage{appendix}
\usepackage{graphicx}

\definecolor{plateblue}{RGB}{31,120,180}
\definecolor{platered}{RGB}{231,41,138}

\DeclareMathOperator*{\argmax}{arg\,max}

\DeclareMathOperator*{\logistic}{\sigma}

\newcommand{\petitioner}{\text{p}}
\newcommand{\respondent}{\text{r}}
\newcommand{\casetitle}[1]{\textit{#1}}

\newcommand{\newcite}[1]{\citeauthor{#1} \shortcite{#1}}
\newcommand{\url}[1]{\texttt{#1}}

\newcommand{\suppBriefTradeoffs}{{\S A}}
\newcommand{\suppLikelihood}{{\S B}}
\newcommand{\suppData}{{\S C}}
\newcommand{\suppTopics}{{\S D}}
\newcommand{\suppJusticeIPs}{{\S E}}
\newcommand{\suppAmiciInfluence}{{\S F}}
\newcommand{\suppUtilityCurve}{{\S G}}

\title{The Utility of Text: The Case of Amicus Briefs and the Supreme Court}
\author{
Yanchuan Sim\\Language Technologies Institute\\Carnegie Mellon University\\Pittsburgh, PA 15213, USA\\\texttt{ysim@cs.cmu.edu} \And
Bryan R. Routledge\\Tepper School of Business\\Carnegie Mellon University\\Pittsburgh, PA 15213, USA\\\texttt{routledge@cmu.edu} \And
Noah A. Smith\\Language Technologies Institute\\Carnegie Mellon University\\Pittsburgh, PA 15213, USA\\\texttt{nasmith@cs.cmu.edu}
}

\begin{document}

\maketitle

\begin{abstract}
\begin{quote}
We explore the idea that authoring a piece of text is an act of maximizing one's expected utility.
To make this idea concrete, we consider the societally important decisions of the Supreme Court of the United States.
Extensive past work in quantitative political science provides a framework for empirically modeling the decisions of justices and how they relate to text.
We incorporate into such a model texts authored by amici curiae (``friends of the court'' separate from the litigants) who seek to weigh in on the decision, then explicitly model their goals in a random utility model.
We demonstrate the benefits of this approach in improved vote prediction and the ability to perform counterfactual analysis.
\end{quote}
\end{abstract}

\section{Introduction}

Some pieces of text are written with clear goals in mind.
Economists and game theorists use the word \emph{utility} for the concept of satisfying a need or desire, and a huge array of theories and models are available for analyzing how utility-seeking agents behave.
This paper takes the first steps in incorporating \emph{text} into these models.

The Supreme Court of the United States (SCOTUS) is the highest court in the American judicial system; its decisions have far-reaching effects.
While the ideological tendencies of SCOTUS' nine justices are widely discussed by press and public, there is a formal mechanism by which organized interest groups can lobby the court on a given case.
These groups are known as \emph{amici curiae} (Latin for ``friends of the court,'' hereafter ``amici,''  singular ``amicus''), and the textual artifacts they author---known as amicus briefs---reveal explicit attempts to sway justices one way or the other.
Taken alongside voting records and other textual artifacts that characterize a case, amicus briefs provide a fascinating setting for empirical study of influence through language.

We build on well-established methodology from political science known as \emph{ideal points} for analyzing votes.
Specifically, \newcite{lauderdale2012scaling} combined descriptive text and ideal points in a probabilistic topic model.
Although the influence of amici has been studied extensively by legal scholars \cite{collins2008friends}, we are the first to incorporate them into ideal points analysis (\S\ref{sec:ip}).
Drawing on decision theory, we then posit amici as rational agents seeking to maximize their expected utility by framing their arguments to influence justices toward a favorable outcome (\S\ref{sec:agents}).
We derive appropriate inference and parameter estimation procedures (\S\ref{sec:learning}).

Our experiments (\S\ref{sec:experiments}) show that the new approach offers substantial gains in vote prediction accuracy.
More importantly, we show how the model can be used to answer questions such as:
How effective were amici on each side of a case?
What would have happened if some or all amicus briefs were not filed?
How might an amicus have changed her brief to obtain a better outcome?
Since our approach characterizes the amicus brief as a (probabilistic) function of the case parameters, our approach could also be used to ask how the amici would have altered their briefs given different merits facts or a different panel of justices.
Although we focus on SCOTUS, our model is applicable to any setting where textual evidence for competing goals is available alongside behavioral response.

\paragraph{SCOTUS Terminology.}
SCOTUS reviews the decisions of lower courts and (less commonly) resolves disputes between states.\footnote{Details about the procedures and rules of the SCOTUS can be found at \url{http://www.uscourts.gov}.}
In a typical case, the \textbf{petitioner} writes a brief putting forward her legal argument; the \textbf{respondent} (the other party) then files a brief.
These, together with a round of responses to each other's initial briefs, are collectively known as \textbf{merits briefs}.
\textbf{Amicus briefs}---further arguments and recommendations on either side---may be filed by groups with an interest (but not a direct stake) in the outcome, with the Court's permission.
After oral arguments (not necessarily allotted for every case) conclude, the justices vote and author one or more opinions.
In this paper, we relate the votes of justices to merits and amicus briefs.

\section{Ideal Point Models} \label{sec:ip}

Ideal point (IP) models are a mainstay in quantitative political science, often applied to voting records to place voters (lawmakers, justices, etc.) in a continuous space.
A justice's ``ideal point'' is a latent variable positioning her in this space.

\paragraph{Unidimensional Ideal Points}  % \label{sec:unidimensional-ip}

The simplest model for judicial votes is a unidimensional IP model \cite{martin2002dynamic}, which posits an IP $\psi_j \in \mathbb{R}$ for each justice $j$.\footnote{\newcite{martin2002dynamic} describe a dynamic unidimensional IP model where justice IPs vary over time. In this work, we assume each justice's IP is fixed over time, for simplicity.}
Often the $\psi_j$ values are interpreted as positions along a liberal-conservative ideological spectrum.
Each case $i$ is represented by \emph{popularity} ($a_i$) and \emph{polarity} ($b_i$) parameters.\footnote{This model is also known as a two parameter logistic model in item response theory literature \cite{fox2010bayesian}, where  $a_i$ is ``difficulty'' and $b_i$ is ``discrimination.''}
A probabilistic view of the unidimensional IP model is that justice $j$ votes in favor of case $i$'s petitioner with probability
\begin{align*}
p(v_{i,j} = \text{petitioner} \mid \psi_j, a_i, b_i) = \sigma\left(a_i + \psi_j b_i\right)
\end{align*}
where $\sigma(x) = \frac{\exp(x)}{1+\exp(x)}$ is the logistic function.
When the popularity parameter $a_i$ is high enough, every justice is more likely to favor the petitioner.
The polarity $b_i$ captures the importance of the justice's ideology $\psi_j$: more polarizing cases (i.e., $|b_i| \gg 0$) push justice $j$ more strongly to the side of the petitioner (if $b_i$ has the same sign as $\psi_j$) or the respondent (otherwise).
While they recover dimensions that maximize statistical fit, IP models conflate many substantive dimensions of opinion and policy, making it difficult to interpret additional dimensions.\footnote{A seminal finding of \newcite{poole1985spatial} is that two dimensions, corresponding to left-right ideology and geographical latitude, explain most of the variance in U.S.~Congressional votes.}
Indeed, such embeddings are ignorant of the issues at stake, or any content of the case, and they cannot generalize to new cases.

\paragraph{Issues and Ideal Points}  % \label{sec:issues-ip}
\newcite{lauderdale2012scaling} incorporate text as evidence and infer dimensions of IP that are grounded in ``topical'' space.
They build on latent Dirichlet allocation (LDA; see \citeauthor{blei2003latent} \citeyear{blei2003latent}), a popular model of latent topics or themes in text corpora.
In their model, each case $i$ is embedded as $\bm\theta_i$ in a $D$-dimensional simplex; the $d$th dimension $\theta_{i,d}$ corresponds to the proportion of case $i$ that is about issue (or, in LDA terminology, topic) $d$.
The probability of justice $j$'s vote is given by
\begin{align*}
p(v_{i,j} = \text{petitioner} \mid  \bm \psi_j, \bm\theta_i, a_i, b_i)
& =  \sigma\left(a_i + \bm\psi_j^\top\left( b_i \bm\theta_i\right)\right)
\end{align*}
where $\psi_{j,d}$ is an \emph{issue-specific} position for justice $j$.
Therefore, the relative degree that each dimension predicts the vote outcome is determined by the text's mixture proportions, resulting in the issue-specific IP $\bm\psi_j^\top\bm\theta_i$.
In their work, they inferred the mixture proportions from justices' opinions, although one can similarly use merits briefs, appeals court opinions, or any other texts that serve as evidence for inferring the issues of a case.

\newcite{lauderdale2012scaling} found that incorporating textual data in this manner\footnote{Of course, LDA is not the only way to ``embed'' a case in a simplex. One can take advantage of expert categorization of case issues. For example, \newcite{gerrish2012howtheyvote} used bill labels as supervision to infer the proportions of issues.} addresses the labeling problem for multidimensional models, and is especially useful for small voting bodies (e.g., SCOTUS), where estimating multidimensional models is difficult due to few observations and variation of preferences across issues.

\paragraph{Amici and Ideal Points}  % \label{sec:amici-ip}
The merits briefs describe the issues and facts of the case.
It is our hypothesis that amicus briefs serve to ``frame'' the facts and, potentially, influence the case outcome.
\newcite{collins2008friends} argued that these organized interest groups play a significant role in shaping justices' choices.
Public interest groups, such as the ACLU and Citizens United, frequently advocate their positions on any case that impinges on their goals.
These briefs can provide valuable assistance to the Court in its deliberation; for example, they can present an argument not found in the merits.\footnote{On occasion, SCOTUS may adopt a position not advanced by either side, but instead urged solely by an amicus brief. Some notable cases are: \emph{Mapp v.~Ohio}, 367 U.S. 643, 646 (1961)
and more recently, \emph{Turner v.~Rogers}, 131 S. Ct. 2507 (2011).}

When filing amicus briefs, amici are required to identify the side they are supporting (or if neither).
However, it is not trivial to automatically tell which side the amici are on as these intentions are not expressed consistently.
We solve this by training a classifier on hand-labeled data (\S\ref{sec:data-preprocessing}).

We propose that amici represent an attempt to shift the position of the case by emphasizing some issues more strongly or framing the case distinctly from the perspectives given in the merits briefs.
The effective position of a case, previously $b_i \bm \theta_i$, is in our model
$b_i \bm \theta_i +  c^\petitioner_i \bm \Delta^\petitioner_i +   c^\respondent_i \bm \Delta^\respondent_i$,
where $c_i^\petitioner$ and $c_i^\respondent$ are the \emph{amicus polarities} for briefs on the side of the petitioner and respondent.
$\bm\Delta^ \petitioner_i$ and $\bm\Delta^\respondent_i$ are the mean issue proportions of the amicus briefs on the side of the petitioner and respondent, respectively.
Our amici-augmented IP model is:
\begin{align}
\lefteqn{p(v_{i,j}  =  \text{petitioner} \mid \psi_j, \bm\theta_i, \bm\Delta_i, a_i, b_i, c_i)} \nonumber
\\  & =  \sigma\left(a_i + \bm\psi_j^\top \left(b_i \bm\theta_i + c^\petitioner_i \bm\Delta^\petitioner_i + c^\respondent_i \bm\Delta^\respondent_i\right)\right)\label{eq:amici-augmented-eqn}
\end{align}
In this model, the vote-specific IP is influenced by two forms of text: legal arguments put forth by the parties involved (merits briefs, embedded in $\bm \theta_i$), and by the amici curiae (amicus briefs, embedded in $\bm \Delta^{\{\petitioner,\respondent\}}_i$), both of which are rescaled independently by the case discrimination parameters to generate the vote probability.
When either $|c^\petitioner_i|$ or $|c^\respondent_i|$ is large (relative to $a_i$ and $b_i$), the vote is determined by the contents of the amicus briefs.
Hereafter, we let $\bm \kappa_i = \langle a_i,b_i, c^\petitioner_i, c^\respondent_i\rangle$.

By letting $\bm\Delta^s_i$ be the average mixture proportions inferred from text of briefs supporting side $s$, we implicitly assume that briefs supporting the same side share a single parameter, and individual briefs on one side influence the vote-specific IP equally.
While \newcite{lynch2004best} and others have argued that some amici are more effective (i.e., influence on justices' votes varies across amicus authors), our model captures the collective effect of amicus briefs and is simple.

\section{Amici as Agents} \label{sec:agents}
In the previous section, the IP models focus on justices' positions embedded in a continuous space.
However, we want to account for the fact that amici are purposeful decision makers who write briefs hoping to sway votes on a case.
Suppose we have an amicus curiae supporting side $s$ (e.g., petitioner), which is presided by a set of justices, $\mathcal{J}$.
The amicus is interested in getting votes in favor of her side, that is, $v_j = s$.
Thus, we assume that she has a simple evaluation function over the outcome of votes $v_1, \dots, v_9$,
\begin{align}
u(v_1,v_2,\dots, v_9) = \textstyle \sum_{j\in\mathcal{J}} \mathbb{I}(v_j=s), \label{eq:utility-function}
\end{align}
where $\mathbb{I}$ is the indicator function.  This is her \textbf{utility}.

\paragraph{Cost of writing.}
In addition to the policy objectives of an amicus, we need to characterize her ``technology'' (or ``budget'') set.
We do this by specifying a cost function, $C$, that is increasing in difference between $\bm\Delta$ and the ``facts'' in $\bm\theta$:
\begin{align*}
C(\bm\Delta, \bm\theta) & = \textstyle\frac{\xi}{2} \|\bm\Delta - \bm\theta\|_2^2
\end{align*}
where $\xi>0$ is a hyperparameter controlling the cost (relative to the vote evaluation).
The function captures the notion that amicus briefs cannot be arbitrary text; there is disutility or effort required to carefully frame a case, or the monetary cost of hiring legal counsel.
The key assumption here is that framing is costly, while simply matching the merits is easy (and presumably unnecessary).
Note the role of the cost function is analogous to regularization in other contexts.

\paragraph*{Expected utility.}
The outcome of the case is uncertain, so the amicus' objective will consider her \emph{expected} utility:\footnote{We use an evaluation function that is linear in votes for simplicity.  The scale of the function is unimportant (expected utility is invariant to affine transformations). However, we leave for future work other specifications of the evaluation function; for example a function that places more emphasis on the majority vote outcome.}
\begin{align*}
\max_{\bm\Delta} \; \mathbb{E}_{\bm\Delta}[u(v_1,\dots, v_9)] - \textstyle\frac{\xi}{2} \|\bm\Delta - \bm\theta\|_2^2
\end{align*}
When an amicus writes her brief, we assume that she has knowledge of the justices' IPs, case parameters, and contents of the merits briefs, but ignores other amici.\footnote{Capturing strategic amici agents (a petitioner amicus choosing brief topics considering a respondent amicus' brief) would require a complicated game theoretical model and, we conjecture, would require a much richer representation of policy and goals.}
As such, taking linearity of expectations, we can compute the expected utility for an amicus on side $s$ using Eq.~\ref{eq:amici-augmented-eqn}:\footnote{The first-order conditions for amicus' purposeful maximization with respect to $\bm\Delta$ lead to interesting brief writing trade offs, which can be found in supplementary \suppBriefTradeoffs.}
\begin{align*}
\max_{\bm\Delta}\;  \textstyle\sum_{j\in\mathcal{J}} \sigma \left(a + \bm\psi_j^\top (b\bm\theta + c^s \bm\Delta)\right)
 - \textstyle \frac{\xi}{2} \|\bm\Delta - \bm\theta\|_2^2 % \label{eq:maximize-expected-utility}
\end{align*}

\paragraph{Random utility models.}
There are several conceivable ways to incorporate amici's optimization into our estimation of justices' IP.
We could maximize the likelihood and impose a constraint that $\bm\Delta$ solve our expected utility optimization (either directly or by checking the first order conditions).\footnote{This is reminiscent of learning frameworks where constraints are placed on the posterior distributions \cite{chang2007guiding,ganchev2010posterior,mccallum2007generalized}. However, the nonlinear nature of our expectations makes it difficult to optimize and characterize the constrained distribution.}
Or, we can view such (soft) constraints as imposing a prior on $\bm\Delta$:
\begin{align}
p_{\text{util}} (\bm\Delta)  \propto \mathbb{E}_{\bm\Delta}[u(v_1,\dots)] + \textstyle\xi(1-\frac{1}{2} \|\bm\Delta - \bm\theta\|_2^2)\label{eq:expected-utility}
\end{align}
where the constant is added so that $p_{\text{util}}$ is non-negative.
Note, were the utility negative, the amici would have chosen not to write a the brief.
This approach is known as  a \textbf{random utility model} in the econometrics discrete-choice literature \cite{mcfadden1974frontiers}.
Random utility models relax the precision of the optimization by assuming that agent preferences also contain an idiosyncratic random component.
Hence, the behavior we observe (i.e., the amicus' topic mixture proportions) has a likelihood that is proportional to expected
 utility.
Considering all the amici, we estimate the full likelihood
\begin{align}
\mathcal{L}(\bm w, \bm v, \bm\psi, \bm\theta, \bm\Delta, \bm \kappa) \times \left[\textstyle\prod_{k\in\mathcal{A}} p_{\text{util}}(\bm\Delta_k) \right]^\eta\label{eq:utility-likelihood}
\end{align}
where $\mathcal{L}(\cdot)$ is the likelihood of our amici IP model (Eq.~\ref{eq:amici-augmented-eqn}), and $\eta$ is a hyperparameter that controls influence of utility on parameter estimation.

Eq.~\ref{eq:utility-likelihood} resembles \emph{product of experts} model \cite{hinton2002training}.
For the likelihood of votes in a case to be maximized, it is necessary that no individual component---generative story for votes, amicus briefs---assigns a low probability.
Accordingly, this results in a principled manner for us to incorporate our assumptions about amici as rational decision makers, each of whom is an ``expert'' with the goal of nudging latent variables to maximize her own expected utility.

\section{Learning and Inference}\label{sec:learning}

\paragraph{Model priors.}
The models we described above combine ideal points, topic models, and random utility; they can be  estimated within a Bayesian framework.
Following \newcite{lauderdale2012scaling}, we place Gaussian priors on the justice and case parameters:
$\rho \sim \mathcal{U}(0, 1)$,
$\bm\psi_j \sim \mathcal{N}(\bm 0, \lambda \mathbf{I} + \rho\bm 1)$ for each justice $j$,
and $\bm\kappa_i \sim \mathcal{N}(\bm 0, \bm\sigma)$ for each case $i$.
The positive off-diagonal elements of the covariance matrix for justice IPs orient the issue-specific dimensions in the same direction (i.e., with conservatives at the same end) and provide shrinkage of IP in each dimension to their common mean across dimensions.
\begin{figure}
\begin{center}
  \includegraphics[width=0.495\linewidth]{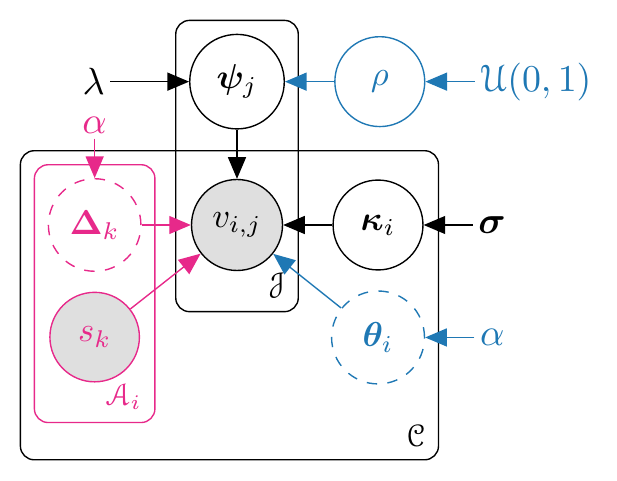}
  \includegraphics[width=0.495\linewidth]{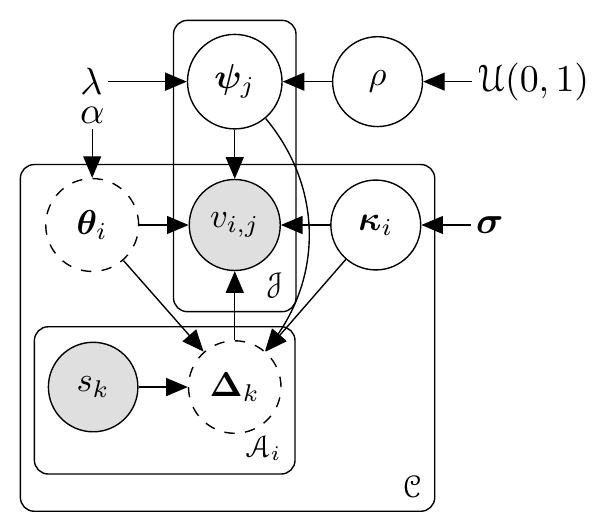}
  \caption{Plate diagrams for the IP models (left) and random utility model (right). $\mathcal{J}$, $\mathcal{C}$ and $\mathcal{A}_i$ are the sets of justices, cases, and amicus briefs (for case $i$), respectively.
$\bm\psi_j$ is the IP for justice $j$;
  $\bm\kappa_i$ is the set of case parameters $a_i, b_i, c^\petitioner_i$ and $c^\respondent_i$ for case $i$; and $\bm\alpha, \bm\sigma, \lambda$, and $\rho$ are hyperparameters.
The mixture proportion nodes (dashed) are fixed in our estimation procedure.
On the left, black nodes comprise the basic IP model, \textcolor{plateblue}{blue} nodes are found in both the issues IP and amici IP models, while \textcolor{platered}{magenta} nodes are found only in the amici IP model.\label{fig:plate-diagram}}
\end{center}
\end{figure}
Fig.~\ref{fig:plate-diagram} presents the plate diagram for the IP models (\S\ref{sec:ip})
on the left and the random utility model (\S\ref{sec:agents}) on the right.

For the non-utility IP models involving text (\S\ref{sec:ip}), LDA is used to infer the latent topic mixtures of text associated with the case and amicus briefs, thus
$\bm\theta_i$ and $\bm\Delta_k$ are both drawn from a symmetric Dirichlet distribution with hyperparameter $\alpha$ and tokens in both sets of texts are drawn from shared topic-word distributions.\footnote{We omit details of LDA, as it is widely known.}

Our random utility model can be described through a similar generative story.
Instead of drawing amicus briefs ($\bm\Delta$) from a Dirichlet, they are drawn from the expected utility distribution (Eq.~\ref{eq:expected-utility}).
The right side of Fig.~\ref{fig:plate-diagram} shows the corresponding plate diagram.
Importantly, note that $\bm\Delta$ here serves as direct evidence for the justice and case parameters, rather than influencing them through v-structures.

\paragraph{Parameter estimation.}
We decoupled the estimation of the topic mixture parameters as a stage separate from the IP parameters.
This approach follows \newcite{lauderdale2012scaling}, who argued for its conceptual simplicity: the text data define the rotation of a multidimensional preference space, while the second stage estimates the locations in that space.
We found in preliminary experiments that similar issue dimensions result from joint vs.~stage-wise inference, but that the latter is much more computationally efficient.

Using LDA,\footnote{We used the C++ implementation of LDA by \newcite{liu2011plda}.} $\bm\theta$ (and, where relevant, $\bm \Delta$) are estimated, then fixed to their posterior means while solving for justice parameters $\bm \psi$ and case parameters $\bm \kappa$.
For the second stage,  we used Metropolis within Gibbs \cite{tierney1994markov}, a hybrid MCMC algorithm, to sample the latent parameters from their posterior distributions.
We sampled $\bm\kappa_i$ for each case and $\bm\psi_j$ for each justice blockwise from a multivariate Gaussian proposal distribution, tuning the diagonal covariance matrix to a target acceptance rate of 15--45\%.
Likewise, $\rho$ is sampled from a univariate Gaussian proposal, with its variance tuned similarly.
For our random utility model, we used the same MCMC approach in sampling the latent IRT variables, but include the expected utility term for each brief in the likelihood function (eq.~\ref{eq:utility-likelihood}).
Details of our sampler and hyperparameter settings can be found in the supplementary materials (\suppLikelihood), while topics and justices' IPs estimated by our model are found in \suppTopics\ and \suppJusticeIPs, respectively.

\paragraph{Data.}\label{sec:data-preprocessing}
We focused on 23 terms of the Court from 1990--2012 \cite{spaeth2013supreme},\footnote{The unit of analysis is the case citation, and we select cases where the type of decision equals 1 (orally argued cases with signed opinions), 5 (cases with equally divided vote), 6 (orally argued per curiam cases), or 7 (judgements of the Court).
In addition, we dropped cases where the winning side was not clear (i.e., coded as ``favorable disposition for petitioning party unclear'').} using texts from LexisNexis.\footnote{\url{http://www.lexisnexis.com}}
We concatenate each of the 2,074 cases' merits briefs from both parties to form a single document, where the text is used to infer the representation of the case in topical space ($\bm\theta$; i.e., merits briefs are treated as ``facts of the case'').
We did not make use of case opinions as did \newcite{lauderdale2012scaling} because opinions are written after votes are cast, tainting the data for predictive modeling.
Each amicus brief is treated as a single document.

As the amicus briefs in our dataset were not explicitly labeled with the side that they support, and manually labeling each brief would be a tedious endeavor, we built a classifier to automatically label the briefs with the side the amici are supporting, taking advantage of cues in the brief content that \emph{strongly} signal the side that the amici is supporting (e.g., ``in support of petitioner'' and ``affirm the judgement'').

Additionally, we find that using only phrases (instead of standard bag-of-words) gave us more interpretable topics \cite{sim2013measuring}.
Details of our phrase extraction, data preprocessing steps, and brief ``side'' classification are in the supplementary materials (\suppData).

\section{Experiments and Analysis}\label{sec:experiments}
\subsection{Vote Prediction}\label{sec:vote-prediction}
We evaluate each model's ability to predict how justices would vote on a case out of the training sample.
To compute the probability of justices' votes, we first infer the topic mixture proportions for the case's merits briefs ($\bm\theta$), and amicus briefs ($\bm\Delta$).
Given all the justice's IPs $\bm\psi_j$, we find the most likely vote outcome for the case by integrating over the case parameters $\bm\kappa$:
\begin{align*}
\argmax_{\bm v}  \textstyle \int_{\bm\kappa} & p(\bm\kappa \mid \bm\sigma)  \textstyle\prod_{j\in\mathcal{J}}  p(v_j \mid \bm\psi_j, \bm\theta, \bm\Delta, \bm\kappa)\\
\times &  \textstyle\prod_{k\in\mathcal{A}} p_{\text{util}}(\bm\Delta_k \mid \bm\psi, \bm\theta, \bm\Delta, \bm\kappa, s_k)
\end{align*}
where $s_k$ is the side brief $k$ supports, and the multiplier is the expected utility term (Eq.~\ref{eq:expected-utility}) which is ignored for the non-utility based models.

Due to the specification of IP models, the probability of a vote is a logistic function of the vote-specific IP, which is a symmetric function implying that justice $j$'s probability of voting towards the petitioner will be the same as if she voted for the respondent when we negate the vote-specific IP.
Thus, we would not be able to distinguish the actual side that the justice will favor, but we can identify the most likely partitioning of the justices into  two groups.\footnote{Given the partitioning of justices, domain experts should be able to identify the side each group of justices would favor.}
We can then evaluate, for each case, an average pairwise accuracy score,
\begin{align*}
\textstyle\binom{9}{2}^{-1} \sum_{j, j'\in\mathcal{J} : j \not = j'} \mathbb{I}[ \mathbb{I}[\hat{v}_j = \hat{v}_{j'}] = \mathbb{I}[v^\ast_j = v^\ast_{j'}]]
\end{align*}
where  $\hat{\bm v}$ ($\bm v^\ast$) are predicted (actual) votes.

We performed 5-fold cross validation, and present the vote partition accuracy in Table~\ref{tab:predict-results}.
\begin{table}
\begin{center}
  \footnotesize
  \begin{tabular}{|l|r|}
    \hline
    \textbf{Model} & \textbf{Accuracy}\\\hline
    Logistic regression w/ topics & 0.715 $\pm$ 0.008\\
    Unanimous & 0.714 $\pm$ 0.003\\
    Unidimensional IP & 0.583 $\pm$ 0.037\\
    Issues IP & 0.671 $\pm$ 0.008\\
    Amici IP & 0.690 $\pm$ 0.021\\
    Random utility IP & \textbf{0.742} $\pm$ 0.006\\\hline
  \end{tabular}
  \caption{Average pairwise vote partition accuracy (five-fold cross-validation).\label{tab:predict-results}}
\end{center}
\end{table}
We have two na\"{i}ve baselines, (i) where all justices vote unanimously, and (ii) where we trained an $\ell_1$-regularized logistic regression classifier for each justice using the concatenated topic proportions of $\bm\theta$ and $\bm\Delta$ as features for each case.
The baselines exhibit similar accuracies, and perform better than when adjusting for issues and/or amici.
This suggests that justices' votes do not always align with their IPs, and that topic models alone may be inadequate for representing IPs.
Furthermore, we believe there may be insufficient information to learn the amicus polarity case parameters ($c^s$) in the amici IP model (which is slightly better than the issues IP model).
However, in the random utility model, amici-agents-experts weigh in, providing additional signals for estimating these parameters, achieving significantly (paired samples $t$-test, $p < 0.001$) better predictive accuracy than  the baseline.

As an additional qualitative validation of our approach, we compared the log-likelihoods between models that consider vs.~ignore amicus briefs.
We found correlations with anecdotal evidence of how justices view the influence of amicus briefs (see \suppAmiciInfluence\ for details).

\subsection{Post Hoc Analysis of Votes}
\begin{figure}[ht]
  \centering
  \includegraphics[width=0.85\linewidth]{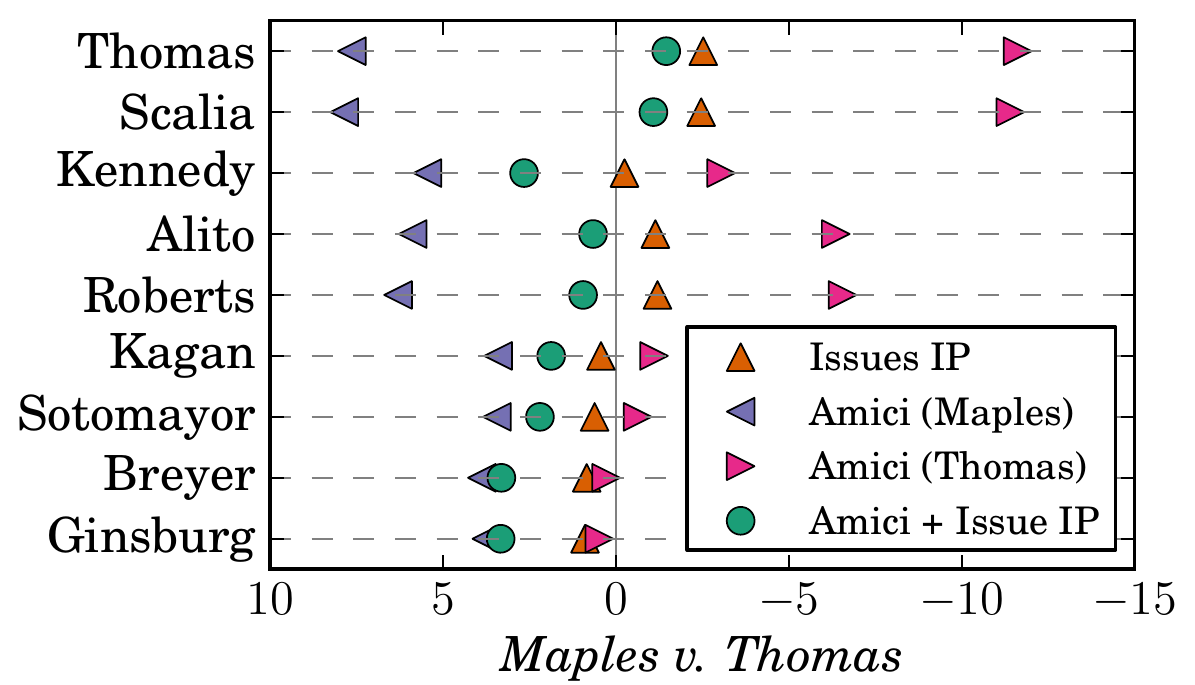}
  \caption{Vote-specific IP estimates decomposed into different influences on each justice's vote on \casetitle{Maples v.~Thomas}. An IP towards the left (right) indicates higher log-odds of vote that is favorable to Maples (Thomas).}
  \label{fig:maples-v-thomas}
\end{figure}

On a case level, we can tease apart the relative contribution each textual component to a justice's decision by analyzing the case parameters learnt by our random utility model.
By zeroing out various case parameters, and plotting them, we can visualize the different impact that each type of text has on a justice's vote-specific IP.
For example, Fig.~\ref{fig:maples-v-thomas} shows the vote-specific IP estimates of justices for the 2011 term death penalty case \casetitle{Maples v.~Thomas} (132 S. Ct. 912).
The issues-only IPs are computed by zeroing out both the amicus polarity parameters ($c^\petitioner$ and $c^\respondent$).
On the other hand, the IP due to amicus briefs supporting Maples (Thomas) is computed by zeroing out only $c^\respondent$ ($c^\petitioner$).
We observe that the issues-only IPs are aligned with each justice's (widely known) ideological stance on the issue of capital punishment.
For instance, the issues-only IPs of Thomas, Scalia, Alito, and Roberts, the strong conservative bloc, favor the respondents (that Maples should not be awarded relief); so did Kennedy, who is widely recognized as the swing justice.
When effects of all amicus briefs are taken into account, the justices' IPs shift toward Maples with varying magnitudes, with the result reflecting the actual ruling (7--2 with Thomas and Scalia dissenting).

\subsection{Counterfactual Analysis}
Following \newcite{pearl2000causality}, we can query the model and perform counterfactual analyses using the vote prediction algorithm (\S\ref{sec:vote-prediction}).
As an illustration, we consider \casetitle{National Federation of Independent Business (NFIB) v. Sebelius (HHS)} (132 S. Ct. 2566), a landmark 2011 case in which the Court upheld Congress's power to enact most provisions of the Affordable Care Act (ACA; ``Obamacare'').\footnote{The case attracted much attention, including a record 136 amicus briefs, of which 76 of these briefs are used in our dataset. 58 (of the 76) were automatically classified as supporting NFIB.}

\begin{figure}[t]
  \begin{center}
    \centering
    \includegraphics[width=0.9\linewidth]{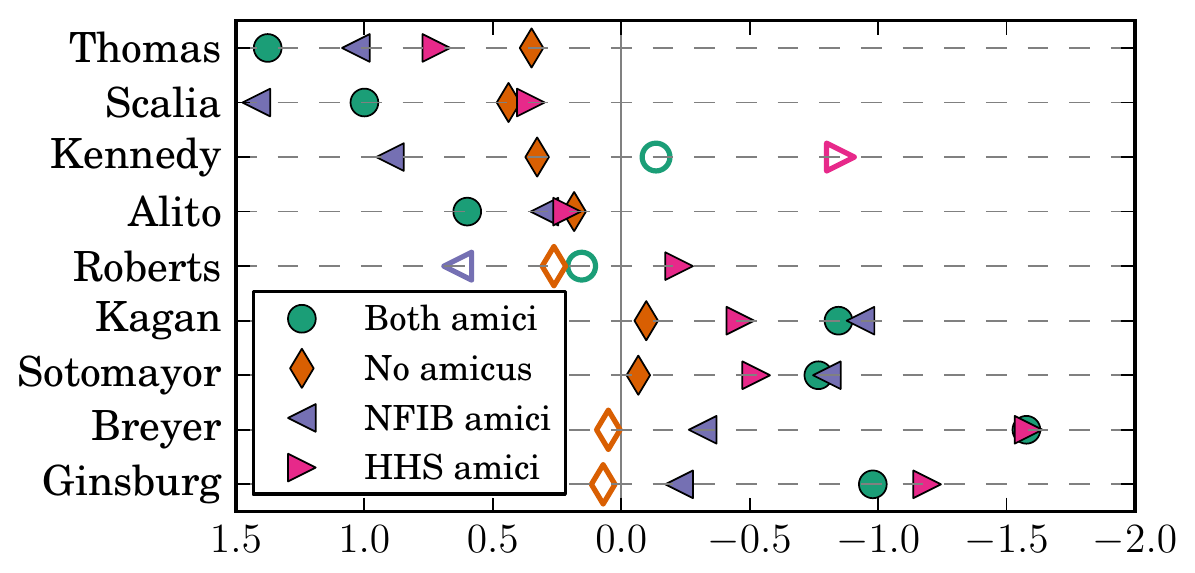}
    % \caption{What if amicus briefs for one side were not filed? \label{fig:business-v-sebelius-only_one_side}}
    \includegraphics[width=0.9\linewidth]{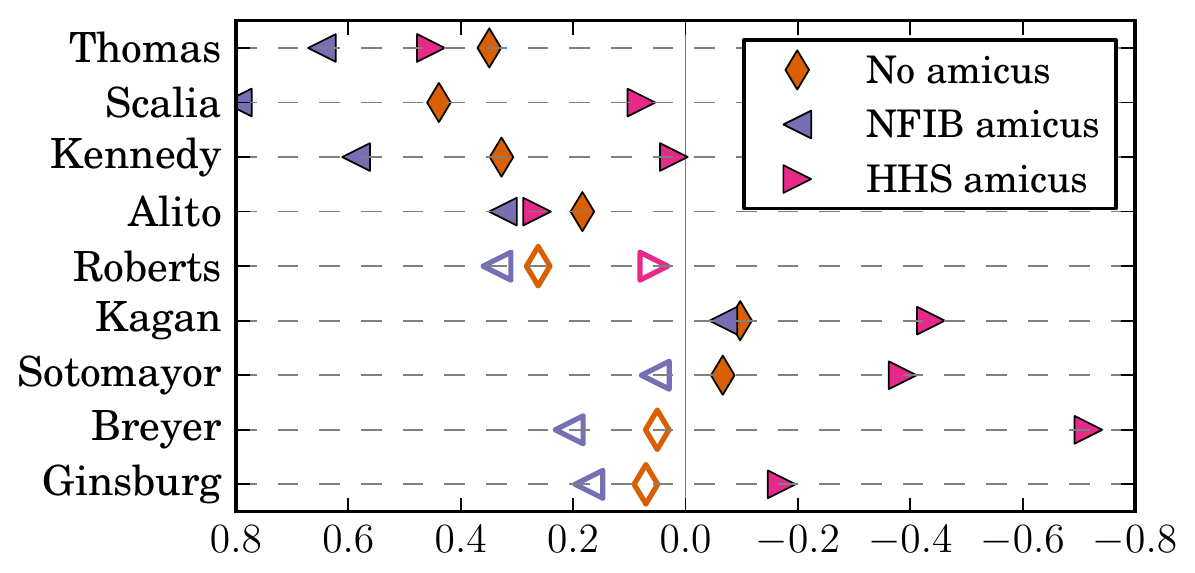}
    % \caption{What if a single amicus files an ``optimally'' written brief? \label{fig:business-v-sebelius-new_amici}}
  \caption{Counterfactual analyses for \casetitle{NFIB v. Sebelius (HHS)}.
  \emph{Top}: What if amicus briefs for one side were not filed?
  \emph{Bottom}: What if a single amicus files an ``optimally'' written brief?
  An IP towards the left (right) indicates higher log-odds of vote favorable to NFIB (HHS). Hollow markers denote that prediction differed from the actual outcome.\label{fig:business-v-sebelius-counterfactuals}}
  \end{center}
\end{figure}

In the merits briefs, the topics discussed revolve around \emph{interstate commerce} and the \emph{individual mandate}, while there is an interesting disparity in topics between briefs supporting NFIB and HHS.\footnote{The merits briefs were estimated at 41\% and 20\%  on the \emph{interstate commerce} and the \emph{individual mandate} topics, respectively.
NFIB amicus briefs were 15\% on \emph{interstate commerce} and 41\% on \emph{individual mandate}; these figures switch to 36\% and 22\% for HHS amicus briefs.}
Notably, amici supporting NFIB are found, on average, to use language concerning \emph{individual mandate}, while amici supporting HHS tend to focus more on topics related to \emph{interstate commerce}.
This is commensurate with the main arguments put forth by the litigants, where NFIB was concerned about the overreach of the government in imposing an individual mandate, while HHS argued that healthcare regulation by Congress falls under the Commerce Clause.
Our model was most uncertain about Roberts and Kennedy, and wrong about both (Fig.~\ref{fig:business-v-sebelius-counterfactuals} top).

\paragraph{Choosing sides.}
The first type of counterfactual analysis that we introduce is, ``What if no (or only one side's) amicus briefs were submitted in the ACA case?''
To answer it, we hold the case out of the training set and attempt to predict the votes under the hypothetical circumstances with the random utility model.
Fig.~\ref{fig:business-v-sebelius-counterfactuals} (top) shows the resulting IP of hypothetical situations where no amicus briefs were filed, or when only briefs supporting one side are filed.
If no amici filed briefs, the model expects that all but Kagan and Sotomayor would favor NFIB, but with uncertainty.
With the inclusion of the amicus briefs supporting NFIB, the model becomes more confident that the conservative bloc of the court would vote in favor of NFIB (except for Alito).  Interestingly, the model anticipates that the same briefs will turn the liberals \emph{away}.
In contrast, the briefs on HHS' side have more success in swaying the case in their favor, especially the crucial swing vote of Kennedy (although it turned out that Kennedy sided with the conservative bloc, and Roberts emerged as the deciding vote in HHS favor).
Consequently, the model can provide insights about judicial decisions, while postulating different hypothetical situations.

\paragraph{Choosing what to write.}
Another counterfactual analysis we can perform, more useful from the viewpoint of the amicus, is, ``how should an amicus frame arguments to best achieve her goals?''
In the context of our model, such an amicus would like to choose the topic mixture $\bm\Delta$ to maximize her expected utility (Eq.~\ref{eq:expected-utility}).
Ideally, one would compute such a topic mixture by maximizing over both $\bm\Delta$ and vote outcome $\bm v$, while integrating over the case parameters.
We resort to a cheaper approximation: analyzing the filer's expected utility curve over two particular topic dimensions: the \emph{individual mandate} and \emph{interstate commerce} topics.
That is, we compute the expected utility curve (see supplementary \suppUtilityCurve) faced by a single amicus as we vary the topic proportions of \emph{individual mandate} and \emph{interstate commerce} topics over multiples of 0.1.

Consequently, the amicus who supports  NFIB can expect to maximize their expected utility (5.2 votes at a cost of 0.21) by ``spending'' about 70\% of their text on \emph{individual mandate}.
On the other hand, the best that an amicus supporting HHS can do is to write a brief that is 80\% about \emph{interstate commerce}, and garner 4.7 votes at a cost of 0.31.
We plot the justices' predicted IPs in Fig.~\ref{fig:business-v-sebelius-counterfactuals} (bottom) using these ``best'' proportions.
The ``best'' proportions IPs are different (sometimes worse) from that in Fig.~\ref{fig:business-v-sebelius-counterfactuals} (top) because in the latter, there are multiple amici influencing the case parameters (through their utility functions) and other topics are present which will sway the justices.
From the perspective of an amicus supporting HHS, the two closest swing votes in the case are Roberts and Kennedy; we know \emph{a posteriori} that Roberts sided with HHS.

\subsection{Discussion}
Our model makes several simplifying assumptions:
(i) it ignores the effects of other amici on a single amicus' writing;
(ii) amici are treated modularly, with a multiplicative effect and no consideration of diminishing returns, temporal ordering, or reputation;
(iii) the cost function does not capture the intricacies of legal writing style (i.e., choice of citations, artful language, etc.);
(iv) the utility function does not fully capture the agenda of each individual amicus;
(v) each amicus brief is treated independently (i.e., no sharing across briefs with the same author), as we do not have access to clean author metadata.
Despite these simplifications, the model is a useful tool for quantitative analysis and hypothesis generation in support of substantive research on the judiciary.

\section{Related Work} \label{sec:related}

\newcite{poole1985spatial} introduced the IP model, using roll call data to infer latent positions of lawmakers.
Since then, many varieties of IP models have been proposed for different voting scenarios: IP models for SCOTUS \cite{martin2002dynamic}, multidimensional IP models for Congressional voting \cite{heckman1996linear,clinton2004statistical}, grounding multidimensional IP models using topics learned from text of Congressional bills \cite{gerrish2012howtheyvote} and SCOTUS opinions \cite{lauderdale2012scaling}.

Amici have been studied extensively, especially their influence on SCOTUS \cite{caldeira1988organized,kearney2000influence,collins2008friends,corley2013influence}.
\newcite{collins2007lobbyists} found that justices can be influenced by persuasive argumentation presented by organized interests.
These studies focus on ideology metadata (liberal/conservative slant of amici, justices, decisions, etc.), disregarding the rich signals encoded in the text of these briefs, whereas we use text as evidence of utility maximizing behavior to study the influence of amicus curiae.

Our model is also related to \newcite{gentzkow2010drives} who model the purposeful ``slant'' of profit-maximizing newspapers looking to gain circulation from consumers with a preference for such slant.
More generally, extensive literature in econometrics estimates structural utility-based decisions \cite[\emph{inter alia}]{berry1995automobile}.

In addition to work on IP models, researchers have used court opinions for authorship \cite{li2013using} and historical analysis, while oral argument transcripts have been used to study power relationships \cite{danescu2012echoes,prabhakaran2013who} and pragmatics \cite{goldwasser2014object}.
% In addition to work on IP models, authorship \cite{li2013using} and historical analysis \cite{wang2012historical} were done using court opinions, while oral argument transcripts have been used to study power relationships \cite{danescu2012echoes,prabhakaran2013who} and pragmatics \cite{goldwasser2014object}.

\section{Conclusion}

We have introduced a random utility model for persuasive text; it is similar to a classical generative model and can be estimated using familiar algorithms.
The key distinction is that persuasive text is modeled as a function of the addressee and the particulars of the matter about which she is being convinced; authors are agents seeking to maximize their expected utility in a given scenario.
In the domain of SCOTUS, this leads to improved vote prediction performance, as the model captures the structure of amicus briefs better than simpler treatments of the text.
Secondly, and more importantly, our model is able to address interesting counterfactual questions.
Were some amicus briefs not filed, or had they been written differently, or had the facts of the case been presented differently, or had different justices presided, our approach can estimate the resulting outcomes.

\section*{Acknowledgements}
This research was supported in part by an A*STAR fellowship to Y.~Sim, NSF grant IIS-1211277, a Google research award, and by computing resources from the Pittsburgh Supercomputing Center.

\appendix
\appendixpage

\section{Brief writing trade offs}

Consider the first-order (or KKT) condition for the purposeful amicus' maximization w.r.t. the $d$th issue:\footnote{We consider inner points for clarity; if $\Delta_d=1$, equality is replaced with $\geq$ (respectively, $0$ and $\leq$).}
\begin{align*}
\lefteqn{ \textstyle\sum_{j\in\mathcal{J}} \sigma' \psi_{j,d} c^s - \xi(\Delta_d - \theta_d)}\nonumber\\
%& \begin{cases}
%  \geq & \text{if } \Delta_d = 1\\
  & = \textstyle\sum_{j\in\mathcal{J}} \sigma' \psi_{j,1} c^s  - \xi(\Delta_1 - \theta_1) & \text{if } 0 < \Delta_d < 1
% \\
 % \leq & \text{if } \Delta_d = 0
%\end{cases}
\end{align*}
where $\sigma'$ is the first order derivative of the vote probability sigmoid function.
Optimality drives all topics (with positive weight) to have equal marginal values.\footnote{Comparing each topic's proportion to topic 1 is arbitrary. In a $D$-topic model, the amicus has only $D-1$ choices.}
The marginal value highlights the tradeoffs an amicus faces, in four components:
(i) the cost of deviating from the merits, i.e., a large difference between $\Delta_d$ and $\theta_d$, is $\xi$;
(ii) a justice whose $\sigma'$ is large, i.e., whose vote  is uncertain, receives greater attention, in particular
(iii) on issues she cares about, i.e., $\psi_{j,d}$ is large, with (iv) $c^s$ controlling the strength of correspondence to issues justices care about.

\section{Likelihood of Random Utility Ideal Point Model}
The likelihood of our amici model with random utility is given below,
\begin{align*}
& p(\bm v, \rho, \bm\psi, \bm a, \bm b, \bm c^{\textnormal{p, r}} \mid \lambda, \bm\theta, \bm\Delta, \bm s)\\
& \propto \prod_{j\in\mathcal{J}} p(\bm\psi_j \mid \lambda, \rho) \prod_{i\in\mathcal{C}} p(a_i, b_i, c^{\text{p, r}}_i \mid \bm\sigma)\\
& \times \prod_{i\in\mathcal{C}} \prod_{j\in\mathcal{J_i}} V_{i,j} \prod_{k\in\mathcal{A}_i} \mathbb{E}[U_{i,k}]^\eta
\end{align*}
where
\begin{align*}
V_{i,j} & = p(v_{i,j} \mid \bm\psi_j, \bm\theta_i, \bm\Delta_i, a_i, b_i, c^{\text{p, r}}_i)\\
& = \logistic \left[(1-v_{i,j}) a_i\right.\\
& \quad\left.+ (1-v_{i,j}) \bm\psi_j^\top (b_i\bm\theta_i + c^\text{p}_i\bm\Delta^\text{p}_i + c_i^\text{r}\bm\Delta^\text{r}_i)\right],
\end{align*}
$\sigma(x) = \frac{\exp(x)}{1+\exp(x)}$ is the logistic function, and
\begin{align*}
\mathbb{E}[U_{i,k}] & = \sum_{j\in\mathcal{J}} \sigma \left(a_i + \bm\psi_j^\top (b_i\bm\theta_i + c^{s_{i,k}}_i \bm\Delta_{i,k})\right)\\
& + \xi\left( 1 - \frac{1}{2} \|\bm\Delta_{i,k} - \bm\theta_i\|_2^2 \right)
\end{align*}
We add a constant to the expected utility term so that it will always be $\geq 0$.
During each iteration of Gibbs sampling, we sampled each latent variable $\rho, \bm\psi_j$ and $[a_i, b_i, c^{\text{p, r}}_i]$ blockwise from the likelihood in turn using the Metropolis-Hastings algorithm.
In each Metropolis-Hastings random walk, we took 500 steps, ignoring the first 250 for burn-in and keeping every 10th step to compute the mean, which we use as our Gibbs update.
In total, we performed 2,000 Gibbs iterations over the training data.

\paragraph*{Hyperparameters.}
We fixed the number of topics in our model to 30.
For LDA, the symmetric Dirichlet parameter for document-topic and topic-word distributions are 0.1 and 0.001, respectively.
For priors on the IP latent variables, we follow the same settings used by \newcite{lauderdale2012scaling}, setting priors on case parameters, $\bm\sigma$, to 4.0, and justice IPs component-wise variance, $\lambda$ to 1.0.
In the random utility model, we set both hyperparameters $\eta$ and $\xi$  to 1.

\section{Data}
We tokenized all the texts and tagged the tokens with the Stanford part of speech tagger \cite{toutanova2003feature}.
We extract $n$-grams with tags that follow the simple (but effective) pattern  {\small \texttt{(Adjective|Cardinal|Noun)+ Noun}}  \cite{justeson1995technical}, representing each document as a ``bag of phrases'', and filtering phrases that appear in less than 25 or more than 3,000 documents obtaining a vocabulary of 55,113 phrase types.
Table~\ref{tab:corpus-statistics} summarizes details of our corpus.
\begin{table}[htp]
  \centering
  \begin{tabular}{|l|r|}
    \hline
    Cases  & 2,074 \\
 \ \ w/amicus briefs & 1,531 \\
     Max.~briefs/case & 76\\
    Word tokens & \multicolumn{1}{|r|}{110.5M} \\  %110,547,378 \\
    Phrase tokens & \multicolumn{1}{|r|}{8.9M} \\ % 8,875,762\\
     Ave.~words/brief  & 3,094 \\
    Ave.~phrases/brief & 339 \\ \hline
  \end{tabular}
  \caption{Corpus statistics ignoring amicus briefs whose supporting side could not be automatically classified confidently.  In the last row, briefs include merits and amicus briefs.\label{tab:corpus-statistics}}
\end{table}

We manually labeled 1,241 randomly selected amicus briefs with its side (petitioner, respondent, neither), and trained a logistic regression classifier\footnote{We used a C++ implementation of logistic regression classifier available at \url{https://github.com/redpony/creg}.} using lexical and formatting features.
We identified 5 sections which are common across almost all briefs and used them as features. The feature templates for our classifier are: $\langle w\rangle, \langle$title,$w\rangle$, $\langle$counsel,$w\rangle$, $\langle$introduction,$w\rangle$, $\langle$statement,$w\rangle$, $\langle$conclusion,$w\rangle$, where $w$ can be any unigram, bigram, or trigram.

We evaluated the performance of our classifier using 5 random splits, with 50\% of our data for training, 30\% for testing, and 20\% for the development set.
We tuned the $\ell_1$-regularization weights on our dev set over the range of coefficients \{0.5, 1, 2, 4, 8, 16\}.
The average accuracy of our classifier is 79.1\%.
Limiting our evaluation to instances whose posterior probability after classification is greater than 0.8, we obtain 90.0\% accuracy and recall of 52.1\%.
Thus, we used 7,258 (out of 13,162) briefs that were classified as supporting neither side or whose posterior probability is $\leq 0.8$ (higher precision at the expense of recall).

\section{Topic Distribution}
Table~\ref{tab:topics} lists the topics and top phrases estimated from our dataset using LDA.
\begin{table*}
  \centering\footnotesize
  \begin{tabularx}{\linewidth}{|c|p{1in}|X|}
    \hline
    \# & \textbf{Topic} & \textbf{Top phrases}\\\hline
    1 & Criminal procedure (1) & reasonable doubt, supervised release, grand jury, prior conviction, plea agreement, controlled substance, guilty plea, double jeopardy clause, sixth amendment, jury trial\\\hline
    2 & Employment & erisa plan, plan administrator, employee benefit plan, insurance company, pension plan, health care, plan participant, individual mandate, fiduciary duty, health insurance\\\hline
    3 & Due process & due process clause, equal protection clause, fundamental right, domestic violence, equal protection, state interest, d e, assisted suicide, controlled substance, rational basis\\\hline
    4 & Indians & m r, m s, indian tribe, tribal court, indian country, fifth amendment, miranda warning, indian affair, vice president, tribal member\\\hline
    5 & Economic activity & attorney fee, limitation period, hobbs act, security law, rule 10b, actual damage, racketeering activity, fiduciary duty, loss causation, security exchange act\\\hline
    6 & Bankruptcy law & bankruptcy court, bankruptcy code, 1996 act, state commission, telecommunication service, network element, eighth circuit, new entrant, pole attachment, communication act\\\hline
    7 & Voting rights & voting right, minority voter, j app, voting right act, covered jurisdiction, fifteenth amendment, redistricting plan, political process, political subdivision, minority group\\\hline
    8 & First amendment & first amendment right, commercial speech, strict scrutiny, cable operator, free speech, first amendment protection, protected speech, child pornography, government interest, public forum\\\hline
    9 & Taxation & interstate commerce, commerce clause, state tax, tax court, gross income, internal revenue code, income tax, dormant commerce clause, state taxation, sale tax\\\hline
    10 & Amicus briefs & national association, amicus brief, vast majority, brief amicus curia, large number, wide range, recent year, public policy, wide variety, washington dc\\\hline
    11 & Labor management & north carolina, collective bargaining agreement, confrontation clause, sta t, north platte river, collective bargaining, inland lake, laramie river, labor organization, re v\\\hline
    12 & Civil action & class action, class member, injunctive relief, final judgment, federal claim, civil action, preliminary injunction, class certification, civil procedure, subject matter jurisdiction\\\hline
    13 & Civil rights & title vii, title vi, civil right act, age discrimination, sexual harassment, old worker, major life activity, reasonable accommodation, prima facie case, disparate impact\\\hline
    14 & State sovereign & sovereign immunity, eleventh amendment, state official, absolute immunity, false claim, private party, 42 usc §1983, state sovereign immunity, eleventh amendment immunity, federal employee\\\hline
    15 & Federal administrations & federal agency, statutory construction, plain meaning, other provision, statutory text, dc circuit, sub §(a), fiscal year, senate report, agency action\\\hline
    16 & Interstate relations & special master, new mexico, prejudgment interest, arkansas river, rt vol, comp act, new jersey, elli island, john martin reservoir, video game\\\hline
    17 & Court of Appeals & eleventh circuit, sixth circuit, circuit court, fourth circuit, oral argument, tenth circuit, further proceeding, appeal decision, instant case, defendant motion\\\hline
    18 & Fourth amendment & fourth amendment, probable cause, arbitration agreement, police officer, national bank, search warrant, exclusionary rule, arbitration clause, reasonable suspicion, law enforcement officer\\\hline
    19 & Eighth amendment & eighth amendment, sex offender, prison official, facto clause, copyright act, copyright owner, unusual punishment, public domain, liberty interest, public safety\\\hline
    20 & International law & international law, foreign state, vienna convention, human right, foreign country, foreign government, jones act, united kingdom, native hawaiian, foreign nation\\\hline
    21 & Equal protection clause & peremptory challenge, law school, equal protection clause, strict scrutiny, high education, racial discrimination, prima facie case, school district, consent decree, compelling interest\\\hline
    22 & Commerce clause (2) & interstate commerce, commerce clause, local government, political subdivision, state regulation, supremacy clause, federal regulation, tobacco product, tenth amendment, federal fund\\\hline
    23 & Immigration law & judicial review, immigration law, final order, removal proceeding, immigration judge, due process clause, deportation proceeding, administrative remedy, compliance order, time limit\\\hline
    24 & Death penalty & death penalty, habeas corpus, reasonable doubt, trial judge, death sentence, ineffective assistance, direct appeal, defense counsel, new rule, mitigating evidence\\\hline
    25 & Environmental issues & navigable water, clean water act, colorado river, special master, project act, public land, fill material, water right, point source, lake mead\\\hline
    26 & Establishment clause & establishment clause, school district, public school, private school, religious school, ten commandment, boy scout, religious belief, religious organization, free exercise clause\\\hline
    27 & Patent law & federal circuit, patent law, prior art, subject matter, expert testimony, lanham act, hazardous substance, patent system, patent act, new drug\\\hline
    28 & Antitrust law & antitrust law, sherman act, contr act, market power, postal service, joint venture, natural gas, high price, public utility, interstate commerce\\\hline
    29 & Election law & political party, taking clause, private property, property owner, fifth amendment, independent expenditure, federal election, property right, contribution limit, general election\\\hline
    30 & Criminal procedure (2) & punitive damage, habeas corpus, second amendment, punitive damage award, enemy combatant, military commission, compensatory damage, state farm, new trial, due process clause\\\hline
  \end{tabularx}
  \caption{Topics and top-10 phrases estimated from briefs using LDA. We manually annotated each topic with the topic labels.}
  \label{tab:topics}
\end{table*}

\section{Justices' Ideal Points}
\begin{sidewaysfigure*}
  \centering
  \includegraphics[width=\linewidth]{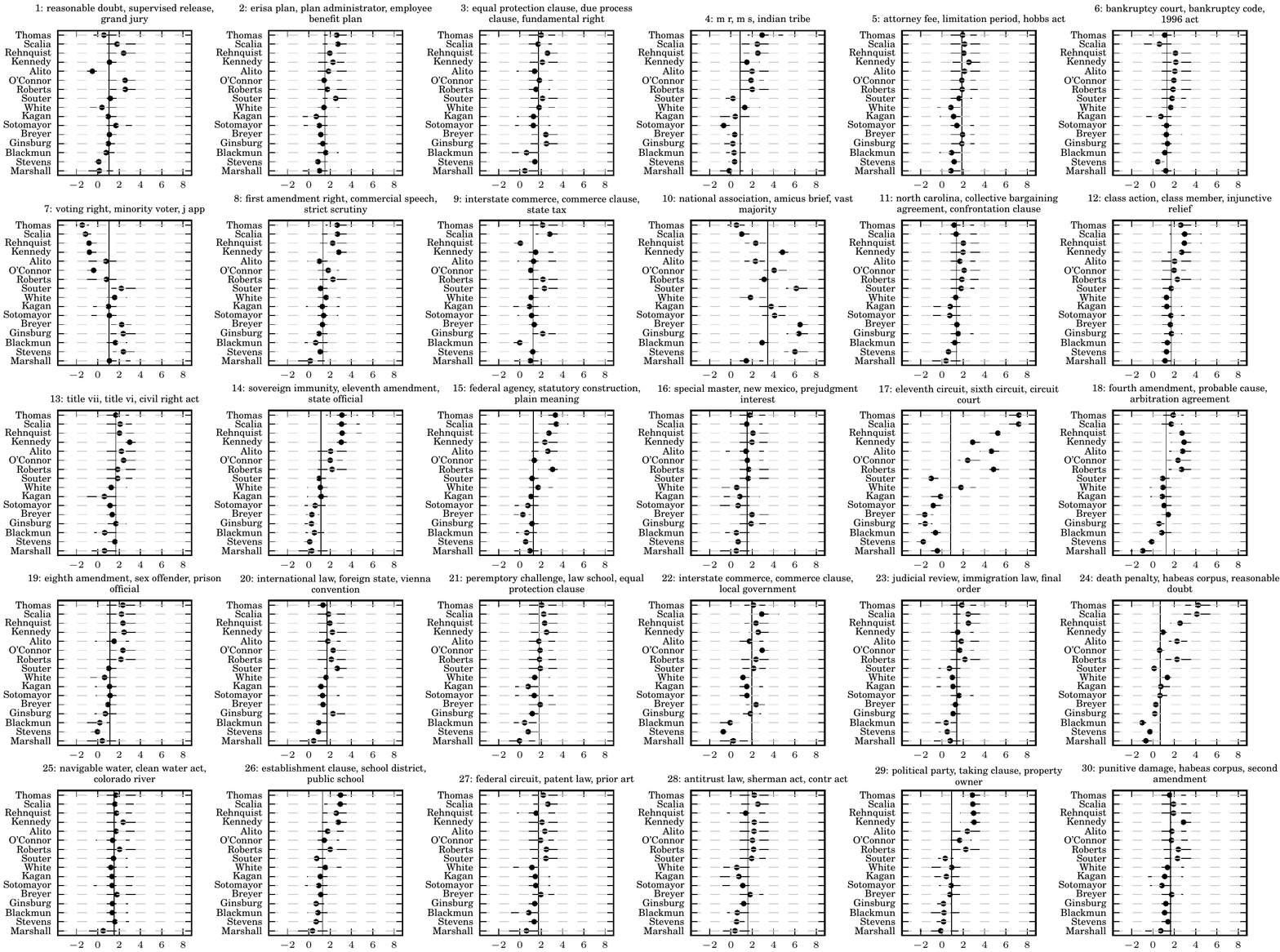}
  \caption{Justices' ideal points by topics. Solid vertical line denotes median ideal point of the justices.}
  \label{fig:justices-ip}
\end{sidewaysfigure*}

The ideal points of justices vary depending on the issues.
We present the justices' ideal points for each of the 30 topics in Fig.~\ref{fig:justices-ip}.

\section{Amici Influence on Justices}
Various justices have expressed opinions on the value of amicus briefs.
Some justices, such as Scalia, are known to be dubious of amicus briefs, preferring to leave the task of reading these briefs to their law clerks, who will pick out any notable briefs for them.\footnote{See \emph{Roper v. Simmons}, 543 U.S. 551, 617-18 (2005) (Scalia J., dissenting); also, ``Don’t re-plow the ground that you expect the parties to plow unless you expect the parties to plow with a particularly dull plow'' (attributed to Scalia on the California Appellate Law Blog, \url{http://bit.ly/1wOmlRE}).}
In contrast, recent surveys of amicus citation rates in justice opinions have often ranked Sotomayor and Ginsburg among justices who most often cite amici in their opinions \cite{franze2011commentary,walsh2013it}.

For each justice, we compute the difference in vote log-likelihood between the issues IP and random utility IP models, as a (noisy) measure of amici influence.
A larger difference in suggests that more of a justice's decision-making can be explained by the presence of amicus briefs.
Table~\ref{tab:justice-influence} presents these differences, and the rankings are consistent with extant hypotheses noted above.
We take this consistency as encouraging, in the spirit of the ``preregistered'' hypotheses of \newcite{sim2013measuring}, but caution that it does not imply a conclusion about causation.
\begin{table}[t]
  \centering\footnotesize
  \begin{tabular}{|l|c|}
    \hline
    \textbf{Justice} & \textbf{Diff.}~$(\times .01)$\\\hline
    John Paul Stevens & 31.5\\
    Clarence Thomas$^\ast$ & 30.8\\
    Thurgood Marshall & 29.4\\
    Elena Kagan$^\ast$ & 28.9\\
    Sonia Sotomayor$^\ast$ & 24.7\\
    John Roberts$^\ast$ & 24.0\\
    Ruth Bader Ginsburg$^\ast$ & 23.9\\
    William Rehnquist & 23.8\\
    Stephen Breyer$^\ast$ & 22.2\\
    David Souter & 22.0\\
    Samuel Alito$^\ast$ & 21.8\\
    Harry Blackmun & 18.0\\
    Byron White & 16.9\\
    Antonin Scalia$^\ast$ & 16.6\\
    Anthony Kennedy$^\ast$ & 15.3\\
    Sandra Day O'Connor & 14.7\\
    \hline
  \end{tabular}
  \caption{Justices ordered by how much amicus briefs can explain their votes, estimated as quadratic mean difference in vote log-likelihood between the issues IP model and the random utility IP model. $^\ast$ denotes current sitting justices.\label{tab:justice-influence}}
\end{table}

\section{Expected utility curve for hypothetical amicus in Obamacare case}
\begin{figure}[ht]
  \centering
  \includegraphics[width=0.8\linewidth]{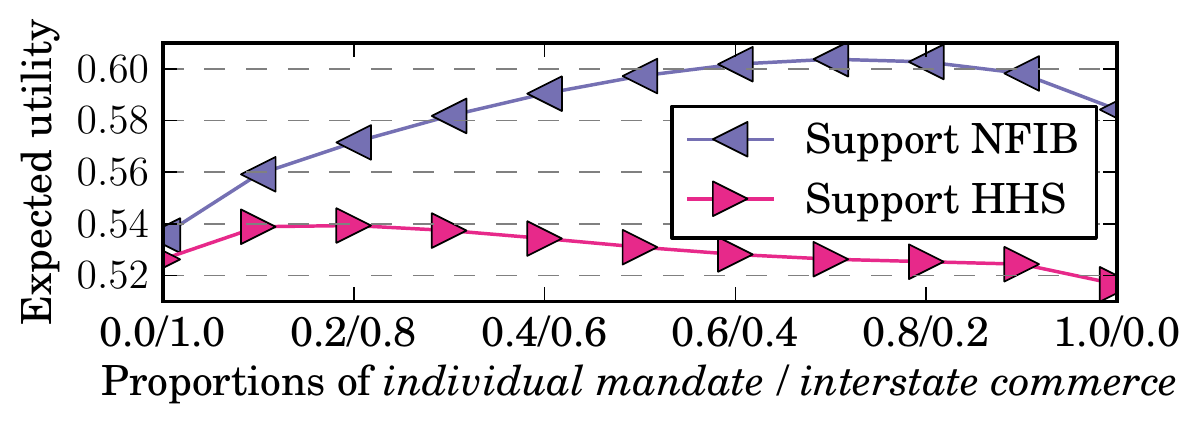}
  \caption{Expected utility when varying between proportions of \emph{individual mandate} and \emph{interstate commerce} topics.}
  \label{fig:business-v-sebelius-utility_curve}
\end{figure}
Fig.~\ref{fig:business-v-sebelius-utility_curve} illustrates the expected utility curve faced by a single amicus as we vary the topic proportions of the \emph{individual mandate} and \emph{interstate commerce} topics.
The model expects an amicus on NFIB's side to get more votes and hence, higher utility, as the model expects justices to be in favor of NFIB prior to amici influence.

We note that interior solutions for the expected utility optimization only exist when proportions are $\in(0, 1)$.
During our experiment, we set the proportions of inactive topics to $10^{-8}$ instead of 0.

\fontsize{9.5pt}{10.5pt}
\selectfont
\bibliographystyle{aaai}
\bibliography{references}

\begin{thebibliography}{}

\bibitem[\protect\citeauthoryear{Berry, Levinsohn, and
  Pakes}{1995}]{berry1995automobile}
Berry, S.; Levinsohn, J.; and Pakes, A.
\newblock 1995.
\newblock Automobile prices in market equilibrium.
\newblock {\em Econometrica: Journal of the Econometric Society}  841--890.

\bibitem[\protect\citeauthoryear{Blei, Ng, and Jordan}{2003}]{blei2003latent}
Blei, D.~M.; Ng, A.~Y.; and Jordan, M.~I.
\newblock 2003.
\newblock Latent {D}irichlet allocation.
\newblock {\em Journal of Machine Learning Research} 3:993--1022.

\bibitem[\protect\citeauthoryear{Caldeira and
  Wright}{1988}]{caldeira1988organized}
Caldeira, G.~A., and Wright, J.~R.
\newblock 1988.
\newblock Organized interests and agenda setting in the {U.S. Supreme Court}.
\newblock {\em American Political Science Review} 82:1109--1127.

\bibitem[\protect\citeauthoryear{Chang, Ratinov, and
  Roth}{2007}]{chang2007guiding}
Chang, M.-W.; Ratinov, L.; and Roth, D.
\newblock 2007.
\newblock Guiding semi-supervision with constraint-driven learning.
\newblock In {\em Proceedings of the 45th Annual Meeting of the Association for
  Computational Linguistics},  280--287.

\bibitem[\protect\citeauthoryear{Clinton, Jackman, and
  Rivers}{2004}]{clinton2004statistical}
Clinton, J.; Jackman, S.; and Rivers, D.
\newblock 2004.
\newblock The statistical analysis of roll call data.
\newblock {\em American Political Science Review} 98:355--370.

\bibitem[\protect\citeauthoryear{Collins}{2007}]{collins2007lobbyists}
Collins, P.~M.
\newblock 2007.
\newblock Lobbyists before the {U.S. Supreme Court}: Investigating the
  influence of amicus curiae briefs.
\newblock {\em Political Research Quarterly} 60(1):55--70.

\bibitem[\protect\citeauthoryear{Collins}{2008}]{collins2008friends}
Collins, P.~M.
\newblock 2008.
\newblock {\em Friends of the {Supreme Court}: Interest Groups and Judicial
  Decision Making}.
\newblock Oxford University Press.

\bibitem[\protect\citeauthoryear{Corley, Collins, and
  Hamner}{2013}]{corley2013influence}
Corley, P.; Collins, P.~M.; and Hamner, J.
\newblock 2013.
\newblock The influence of amicus curiae briefs on {U.S.~Supreme Court} opinion
  content.
\newblock In {\em APSA 2013 Annual Meeting Paper}.

\bibitem[\protect\citeauthoryear{Danescu-Niculescu-Mizil \bgroup et
  al\mbox.\egroup }{2012}]{danescu2012echoes}
Danescu-Niculescu-Mizil, C.; Lee, L.; Pang, B.; and Kleinberg, J.
\newblock 2012.
\newblock Echoes of power: {Language} effects and power differences in social
  interaction.
\newblock In {\em Proceedings of the 21st International Conference on World
  Wide Web}, WWW '12,  699--708.

\bibitem[\protect\citeauthoryear{Fox}{2010}]{fox2010bayesian}
Fox, J.-P.
\newblock 2010.
\newblock {\em {Bayesian Item Response Modeling: Theory and Applications}}.
\newblock Statistics for Social and Behavioral Sciences. Springer.

\bibitem[\protect\citeauthoryear{Franze and
  Anderson}{2011}]{franze2011commentary}
Franze, A.~J., and Anderson, R.~R.
\newblock 2011.
\newblock Commentary: The {Court}'s increasing reliance on amicus curiae in the
  past term.
\newblock \url{http://www.arnoldporter.com/resources/
  documents/Arnold\&PorterLLP\_NationalLaw Journal\_8.24.11.pdf}.
\newblock Accessed: 08-25-2014.

\bibitem[\protect\citeauthoryear{Ganchev \bgroup et al\mbox.\egroup
  }{2010}]{ganchev2010posterior}
Ganchev, K.; Gra\c{c}a, J.~a.; Gillenwater, J.; and Taskar, B.
\newblock 2010.
\newblock Posterior regularization for structured latent variable models.
\newblock {\em Journal of Machine Learning Research} 11:2001--2049.

\bibitem[\protect\citeauthoryear{Gentzkow and
  Shapiro}{2010}]{gentzkow2010drives}
Gentzkow, M., and Shapiro, J.~M.
\newblock 2010.
\newblock What drives media slant? evidence from {U.S.}~daily newspapers.
\newblock {\em Econometrica} 78(1):35--71.

\bibitem[\protect\citeauthoryear{Gerrish and
  Blei}{2012}]{gerrish2012howtheyvote}
Gerrish, S., and Blei, D.~M.
\newblock 2012.
\newblock How they vote: Issue-adjusted models of legislative behavior.
\newblock In Pereira, F.; Burges, C.; Bottou, L.; and Weinberger, K., eds.,
  {\em Advances in Neural Information Processing Systems 25},  2753--2761.
\newblock Curran Associates, Inc.

\bibitem[\protect\citeauthoryear{Goldwasser and
  Daum{\'e}~III}{2014}]{goldwasser2014object}
Goldwasser, D., and Daum{\'e}~III, H.
\newblock 2014.
\newblock ``{I} object!'' modeling latent pragmatic effects in courtroom
  dialogues.
\newblock In {\em Proceedings of the 14th Conference of the European Chapter of
  the Association for Computational Linguistics},  655--663.

\bibitem[\protect\citeauthoryear{Heckman and Snyder}{1996}]{heckman1996linear}
Heckman, J.~J., and Snyder, Jr., J.~M.
\newblock 1996.
\newblock Linear probability models of the demand for attributes with an
  empirical application to estimating the preferences of legislators.
\newblock Working Paper 5785, National Bureau of Economic Research.

\bibitem[\protect\citeauthoryear{Hinton}{2002}]{hinton2002training}
Hinton, G.~E.
\newblock 2002.
\newblock Training products of experts by minimizing contrastive divergence.
\newblock {\em Neural Computation} 14(8):1771--1800.

\bibitem[\protect\citeauthoryear{Justeson and
  Katz}{1995}]{justeson1995technical}
Justeson, J.~S., and Katz, S.~M.
\newblock 1995.
\newblock Technical terminology: Some linguistic properties and an algorithm
  for identification in text.
\newblock {\em Natural Language Engineering} 1:9--27.

\bibitem[\protect\citeauthoryear{Kearney and
  Merrill}{2000}]{kearney2000influence}
Kearney, J.~D., and Merrill, T.~W.
\newblock 2000.
\newblock The influence of amicus curiae briefs on the {Supreme Court}.
\newblock {\em University of Pennsylvania Law Review}  743--855.

\bibitem[\protect\citeauthoryear{Lauderdale and
  Clark}{2014}]{lauderdale2012scaling}
Lauderdale, B.~E., and Clark, T.~S.
\newblock 2014.
\newblock Scaling politically meaningful dimensions using texts and votes.
\newblock {\em American Journal of Political Science} 58(3):754--771.

\bibitem[\protect\citeauthoryear{Li \bgroup et al\mbox.\egroup
  }{2013}]{li2013using}
Li, W.; Azar, P.; Larochelle, D.; Hill, P.; Cox, J.; Berwick, R.~C.; and Lo,
  A.~W.
\newblock 2013.
\newblock Using algorithmic attribution techniques to determine authorship in
  unsigned judicial opinions.
\newblock {\em Stanford Technology Law Review}  503--534.

\bibitem[\protect\citeauthoryear{Liu \bgroup et al\mbox.\egroup
  }{2011}]{liu2011plda}
Liu, Z.; Zhang, Y.; Chang, E.~Y.; and Sun, M.
\newblock 2011.
\newblock {PLDA}+: Parallel latent {D}irichlet allocation with data placement
  and pipeline processing.
\newblock {\em ACM Transactions on Intelligent Systems and Technology}.
\newblock Software available at \url{http://code.google.com/p/plda}.

\bibitem[\protect\citeauthoryear{Lynch}{2004}]{lynch2004best}
Lynch, K.~J.
\newblock 2004.
\newblock Best friends -- {Supreme Court} law clerks on effective amicus curiae
  briefs.
\newblock {\em Journal of Law \& Politics} 20:33.

\bibitem[\protect\citeauthoryear{Martin and Quinn}{2002}]{martin2002dynamic}
Martin, A.~D., and Quinn, K.~M.
\newblock 2002.
\newblock Dynamic ideal point estimation via {Markov chain Monte Carlo} for the
  {U.S.~Supreme Court}, 1953--1999.
\newblock {\em Political Analysis} 10(2):134--153.

\bibitem[\protect\citeauthoryear{McCallum, Mann, and
  Druck}{2007}]{mccallum2007generalized}
McCallum, A.; Mann, G.; and Druck, G.
\newblock 2007.
\newblock Generalized expectation criteria.
\newblock Technical Report UM-CS-2007-60, University of Massachusetts, Amherst,
  MA 01003, USA.

\bibitem[\protect\citeauthoryear{McFadden}{1974}]{mcfadden1974frontiers}
McFadden, D.
\newblock 1974.
\newblock Conditional logit analysis of qualitative choice behavior.
\newblock In Zarembka, P., ed., {\em Frontiers in Econometrics}. New York:
  Academic Press.
\newblock  105--142.

\bibitem[\protect\citeauthoryear{Pearl}{2000}]{pearl2000causality}
Pearl, J.
\newblock 2000.
\newblock {\em {Causality: Models, Reasoning, and Inference}}.
\newblock Cambridge University Press.

\bibitem[\protect\citeauthoryear{Poole and Rosenthal}{1985}]{poole1985spatial}
Poole, K.~T., and Rosenthal, H.
\newblock 1985.
\newblock A spatial model for legislative roll call analysis.
\newblock {\em American Journal of Political Science} 29(2):357--384.

\bibitem[\protect\citeauthoryear{Prabhakaran, John, and
  Seligmann}{2013}]{prabhakaran2013who}
Prabhakaran, V.; John, A.; and Seligmann, D.~D.
\newblock 2013.
\newblock Who had the upper hand? ranking participants of interactions based on
  their relative power.
\newblock In {\em Proceedings of the 6th International Joint Conference on
  Natural Language Processing},  365--373.

\bibitem[\protect\citeauthoryear{Sim \bgroup et al\mbox.\egroup
  }{2013}]{sim2013measuring}
Sim, Y.; Acree, B. D.~L.; Gross, J.~H.; and Smith, N.~A.
\newblock 2013.
\newblock Measuring ideological proportions in political speeches.
\newblock In {\em Proceedings of the Conference on Empirical Methods in Natural
  Language Processing},  91--101.

\bibitem[\protect\citeauthoryear{Spaeth \bgroup et al\mbox.\egroup
  }{2013}]{spaeth2013supreme}
Spaeth, H.~J.; Benesh, S.; Epstein, L.; Martin, A.~D.; Segal, J.~A.; and Ruger,
  T.~J.
\newblock 2013.
\newblock {Supreme Court} database, version 2013 release 01.
\newblock Database at \url{http://supremecourtdatabase.org}.

\bibitem[\protect\citeauthoryear{Tierney}{1994}]{tierney1994markov}
Tierney, L.
\newblock 1994.
\newblock {Markov} chains for exploring posterior distributions.
\newblock {\em The Annals of Statistics} 22(4):1701--1728.

\bibitem[\protect\citeauthoryear{Toutanova \bgroup et al\mbox.\egroup
  }{2003}]{toutanova2003feature}
Toutanova, K.; Klein, D.; Manning, C.~D.; and Singer, Y.
\newblock 2003.
\newblock Feature-rich part-of-speech tagging with a cyclic dependency network.
\newblock In {\em Proceedings of NAACL-HLT},  173--180.

\bibitem[\protect\citeauthoryear{Walsh}{2013}]{walsh2013it}
Walsh, M.
\newblock 2013.
\newblock {Supreme Court} report: It was another big term for amicus curiae
  briefs at the high court.
\newblock \url{http://www.abajournal.com/magazine/arti
  cle/it\_was\_another\_big\_term\_for\_amicus\_cur
  iae\_briefs\_at\_the\_high\_court}.
\newblock Accessed: 08-25-2014.

\end{thebibliography}

\end{document}